\documentclass[runningheads]{llncs}

\usepackage{stmaryrd, url}

\usepackage{fancyvrb}
\usepackage{amsmath}
\usepackage{amssymb}  \usepackage{graphicx}
\usepackage[all]{xy}
\usepackage{longtable}

\usepackage{url}
\urldef{\mailsa}\path|{gchu,pjs}@csse.unimelb.edu.au|

\ifx\pdftexversion\undefined
  \usepackage[dvips]{color}
\else
  \usepackage[pdftex]{color}
\fi

\definecolor{grey}{rgb}{0.6,0.6,0.6}

\definecolor{lightgrey}{rgb}{0.92,0.92,0.92}

\newcommand{\ignore}[1]{}
\newcommand{\short}[1]{}

\newlength{\onwidth}
\newcommand{\alldiff}{\mathit{alldiff}}

\newcommand{\tablec}{\mathit{table}}

\newcommand{\circuit}{\mathit{circuit}}
\newcommand{\disjunctive} {\mathit{disjunctive}}

\newcommand{\chuffed}{\textsc{Chuffed}}

\newcommand{\sched}[2]{#1 \lhd #2}
\renewcommand{\sched}[2]{s_{#1} + d_{#1} \leq s_{#2}}
\renewcommand{\sched}[2]{#1 \ll #2}

\newcommand{\essential}[1]{\left \llbracket #1 \right \rrbracket}
\newcommand{\lit}[1]{\essential{#1}}

\newcommand{\obj} {\mathit{f}}

\newcommand{\pjs}[1]{\marginpar{\sc Pjs}\textbf{\color{blue}[#1]}}
\newcommand{\gchu}[1]{\marginpar{\sc gchu}\textbf{\color{red}[#1]}}

     {\end{tabular}\end{tt}\end{small}}

\newcommand{\true}{\mathit{true}}
\newcommand{\false}{\mathit{false}}

\newcommand{\nolcg}[1]{}

\begin{document}

\mainmatter 

\title{Structure Based Extended Resolution for Constraint Programming}
\titlerunning{Structure Based Extended Resolution for Constraint Programming}

%
%
\author{Geoffrey Chu\inst{2} \and Peter
  J. Stuckey\inst{1,2} }
\authorrunning{Chu and Stuckey}

\institute{National ICT Australia, Victoria Laboratory
\and
  Department of Computing and Information Systems, \\
  University of Melbourne, Australia \\
  \mailsa}
%
%

\maketitle

\begin{abstract}

  Nogood learning is a powerful approach to reducing search in Constraint
  Programming (CP) solvers. The current state of the art, called Lazy Clause
  Generation (LCG), uses resolution to derive nogoods expressing the reasons
  for each search failure. Such nogoods can prune other parts of the search
  tree, producing exponential speedups on a wide variety of problems. Nogood
  learning solvers can be seen as resolution proof systems. The stronger the
  proof system, the faster it can solve a CP problem. It has recently been
  shown that the proof system used in LCG is at least as strong as general
  resolution. However, stronger proof systems such as \emph{extended
    resolution} exist. Extended resolution allows for literals expressing
  arbitrary logical concepts over existing variables to be introduced and
  can allow exponentially smaller proofs than general resolution. The
  primary problem in using extended resolution is to figure out exactly
  which literals are useful to introduce. In this paper, we show that we can
  use the structural information contained in a CP model in order to
  introduce useful literals, and that this can translate into significant
  speedups on a range of problems.

\end{abstract}

\section{Introduction}
\label{sec:intro}

Nogood learning is a powerful approach to reducing search in Constraint
Programming (CP) solvers. The current state of the art is Lazy Clause
Generation~\cite{Ohrimenko2009} (LCG). LCG adapts the clause learning
techniques from Boolean Satisfiability (SAT) to the more generic domain of
CP problems where we have finite domain variables and global
constraints. Each propagator in an LCG solver is instrumented so that it is
able to explain each of its propagations using a clause. These clauses form
an implication graph. When a search failure occurs, the implication graph is
analyzed and resolution is performed on the clauses in the implication graph
in order to derive a nogood which explains the reasons for the
failure. These nogoods can then be propagated in order to prune other parts
of the search tree. Nogood learning is very effective on structured problems
and can often provide orders of magnitude speedup over a non-learning
solver.

A complete search CP solver can be seen as a proof system which is trying to prove that no solution exists in a satisfiability problem, or that no solution better than a certain objective value exists in an optimization problem. It either finds a counter example (i.e., a solution) during the proof process, or it succeeds in proving that no solution exists. The size of the search tree is bounded from below by the size of the smallest proof possible in the proof system. Thus in general, the stronger the proof system used, the faster a CP solver can solve the problem. The proof system used in an LCG solver is much stronger than the one used in a non-learning solver. This is because when a non-learning CP solver fails a subtree, it has only proved that that particular subtree fails. On the other hand, when a LCG solver fails a subtree, it uses resolution to derive a nogood that proves that this subtree fails, but also that other similar subtrees (i.e., those which satisfies the conditions in the nogood) also fail.

It has recently been proved~\cite{darwiche} 
that a SAT solver performing conflict directed
clause learning and restarts has a proof system that is as powerful as
\emph{general resolution}~\cite{resolution}. The proof system of LCG
solvers, which inherits the resolution based learning of SAT solvers and the possibly non-resolution based inferences of CP's global propagators, is even more powerful. However, the power of the resolution part of the proof system is constrained by the
set of literals that it is allowed to use in the proof. We call this set of
literals the \emph{language} of the resolution proof system. \emph{Extended
  resolution}~\cite{er} is a proof system even stronger than general
resolution and is one of the most powerful proof systems for propositional
logic~\cite{Urquhart98}. It allows the language of the resolution proof
system to be dynamically extended during runtime by introducing new
variables representing arbitrary logical expressions over the existing
ones. It is well known that there can be an exponential separation between
the size of the proof generated by extended resolution and general
resolution on certain problems~\cite{pigeonhole}. Clearly, utilizing
extended resolution in nogood learning could be a very effective way to
improve the speed of a CP solver.

While extended resolution offers the potential for significant speedups,
such speedups are often difficult to realize in practice. This is because in
order for extended resolution to produce shorter proofs, it is necessary for
the system to make the right \emph{extensions}, i.e., introduce variables
expressing the right logical concepts. It is typically very difficult to
know which particular extensions are needed to speed up a proof, so it is
difficult to use extended resolution effectively. There have been several
attempts at augmenting SAT solvers with extended resolution capabilities
with varying success (e.g.,~\cite{audemard10,huang10}). Constraint Programming provides a unique opportunity for the
effective usage of extended resolution. In Constraint Programming, problems
are modeled in terms of high level constraints which preserve much of the
structure of the problem. This structural information provides important
information regarding which extensions will be useful, and allows us to
exploit the potential speedups made possible by extended resolution.

Most of the major advances in resolution based nogood learning in CP has
come about due to an extension of the language of the resolution proof
system in order to exploit the structure of finite domain integer
variables. The earlier works on 
nogood learning~(see e.g. \cite{dechter}, chapter 6) 
only considers \emph{equality literals} of the form $x = v$ where $x$ is a
variable and $v$ is a value. Later on, \emph{disequality literals} of the
form $x \neq v$ were introduced~\cite{bachus,gnogoods}. 
More recently, \emph{inequality literals} of the form $x \geq v, x \leq v$
were introduced in LCG~\cite{Ohrimenko2009}. Each extension significantly increased the expressiveness of the nogoods and the power of the proof system, resulting in significant speedups compare to previous versions. In this paper, we look at other types of structure that can be found in CP problems and consider how they can be exploited via the introduction of additional literals into the language. Our main contributions are as follows:

\begin{itemize}
\item We provide a framework for assessing the generality of an explanation generated by an LCG propagator, given a fixed language of resolution $L$.
\item We examine the internal structure of commonly used global constraints to see which kinds of language extensions can be useful for improving the power of the resolution proof system.
\item We show that the global structure of the problem can be used in order to decide which language extension to make.
\end{itemize}

\ignore{
In the next section we define our terminology and briefly review constraint
programming and lazy clause generation.
In Section~\ref{sec:gen} we discuss the generality of explanations and how
this is affected by the language of resolution.
In Section~\ref{sec:ext-lang} we show the benefit of extending the language
of resolution and then examine common global constraints to determine
potential extensions. 
In Section~\ref{sec:struct} we discuss how to take advantage of the global
structure of the problem in choosing language extensions for the linear constraint.
In Section~\ref{sec:exp} we give experiments showing the benefits of a
stronger language of resolution.
In Section~\ref{sec:related} we discuss related work and finally
in Section~\ref{sec:conc} we conclude.
}

\section{Definitions and Background}
\label{sec:def}

Let $\equiv$ denote syntactic identity, $\Rightarrow$ denote logical implication and $\Leftrightarrow$ denote logical equivalence. A \emph{constraint satisfaction problem} (CSP) is a tuple $P \equiv (V, D, C)$, where $V$ is a set of variables, $D$ is a set of (unary) domain constraints, and $C$ is a set of (n-ary) constraints. An assignment $\theta$ is a \emph{solution} of $P$ if it satisfies every constraint in $D$ and $C$. In an abuse of notation, if a symbol $C$ refers to a set of constraints $\{c_1, \ldots, c_n\}$, we will often also use the symbol $C$ to refer to the conjunction $c_1 \wedge \ldots \wedge c_n$.

CP solvers solve CSP's by interleaving search with inference. We begin with the original problem at the root of the search tree. At each node in the search tree, we propagate the constraints to try to infer variable/value pairs which can no longer be taken in any solution in this subtree. Such pairs are removed from the current domain. If some variable's domain becomes empty, then the subtree has no solution and the solver backtracks. If all the variables are assigned and no constraint is violated, then a solution has been found and the solver can terminate. If inference is unable to detect either of the above two cases, the solver further divides the problem into a number of more constrained subproblems and searches each of those in turn.

A CP solver implementing LCG has a number of additional features which allow
it to perform nogood learning. Firstly, for each integer variable $x$ with
initial domain $\{l, \ldots, u\}$, the solver adds Boolean variables to
represent the truth value of the logical expressions $x = v$ for $v = l,
\ldots, u$ and $x \geq v$ for $v = l+1, \ldots, u$. We use $\lit{e}$ to
denote the Boolean variable which represents the truth value of logical
expression $e$. The solver enforces the 
channeling constraint $\lit{e} \leftrightarrow e$ for each such variable. So
for example, the Boolean variable $\lit{x = 5}$ is true iff $x
= 5$ is implied by the current domain $D$. For convenience, we also use $\lit{x \neq v}$ to refer to $\neg\lit{x = v}$ and $\lit{x \leq v}$ to refer to $\neg\lit{x \geq v+1}$. We call literals of form $\lit{x = v}$ equality literals, literals of form $\lit{x \neq v}$ disequality literals, and literals of form $\lit{x \geq v}$ or $\lit{x \leq v}$ inequality literals.

In an LCG solver (and indeed most CP solvers), the only allowed kinds of domain changes are: fixing a variable to a value, removing a value, increasing the lower bound, or decreasing the upper bound. Each of these can be expressed as setting one of the literals $\lit{x = v}$, $\lit{x \neq v}$, $\lit{x \geq v}$ or $\lit{x \leq v}$ true. Each propagator in an LCG solver is instrumented in order to explain each of its domain changes with a clause called the \emph{explanation}.

\begin{definition}
Given current domain $D$, suppose the propagator for constraint $c$ makes an inference $p$, i.e., $c \wedge D \Rightarrow p$. An \emph{explanation} for this inference is a clause: $\mathit{expl}(p) \equiv l_1 \wedge \ldots \wedge l_k \rightarrow p$ where $l_i$ and $p$ are literals, s.t. $c \Rightarrow \mathit{expl}(p)$ and $D \Rightarrow l_1 \wedge \ldots \wedge l_k$.
\end{definition}

For example, given constraint $x \leq y$ and current domain $x \in \{3,4,5\}$, the propagator may infer that $y \geq 3$, with the explanation $\lit{x \geq 3} \rightarrow \lit{y \geq 3}$. The explanation $expl(p)$ explains why $p$ has to hold given $c$ and the current domain $D$. We can consider $expl(p)$ as the fragment of the constraint $c$ from which we inferred that $p$ has to hold. We call the set of literals available for forming explanations the \emph{language of resolution} for the system.

As propagation proceeds, these explanations form an acyclic implication graph. Whenever a conflict is found by an LCG solver, the implication graph can be analyzed in order to derive a set of sufficient conditions for the conflict to reoccur. Just as most current state of the art SAT solvers, LCG solvers derive the \emph{first unique implication point} (1UIP) nogood. This is done by repeatedly resolving the conflicting clause (the clause explaining the conflict) with the explanation clause for the latest inferred literal until the clause contains only one literal from the current decision level. The resulting clause, or nogood as it is more commonly called in CP, is an implied constraint of the problem which proves that this particular subtree failed. However, this nogood often also proves that other subtrees fail for a similar reason to the current one. Thus we can add the nogood as a propagator to prune other parts of the search tree.

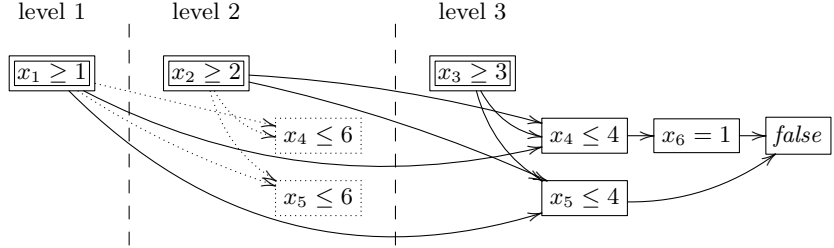
\begin{figure}[t]
\footnotesize 
\newcommand{\xyo}[1]{*+[F-]{#1}}
\newcommand{\xyd}[1]{*+[F=]{#1}} 
\newcommand{\xydd}[1]{*+[F.]{#1}}
$$
\xymatrix@=10pt{ 
   && \mbox{level 1} & \ar@{--}[dddd] & \mbox{level 2} && \ar@{--}[dddd]
& \mbox{level 3} \\
  && \xyd{x_1 \geq 1}
   \ar@{.>}@/_0pt/[drrr]\ar@{.>}@/_5pt/[ddrrr]
   \ar@{->}@/_22pt/[drrrrrr]\ar@{->}@/_35pt/[ddrrrrrr] 
   &&   \xyd{x_2 \geq 2}
   \ar@{.>}@/_10pt/[dr]\ar@{.>}@/_10pt/[ddr]
   \ar@{->}@/^5pt/[drrrr]\ar@{->}@/^5pt/[ddrrrr]
  &&& \xyd{x_3 \geq 3}
      \ar@{->}@/_10pt/[dr] 
      \ar@{->}@/_10pt/[ddr]
 \\  
  &&&&& \xydd{x_4 \leq 6} &&& \xyo{x_4 \leq 4} \ar[r] & \xyo{x_6 = 1}\ar[r] & \xyo{\false} \\
  &&&&& \xydd{x_5 \leq 6} &&& \xyo{x_5 \leq 4} \ar@/_10pt/[rru] &   \\
&&&&&&&&&&&&&&&&& \\
}
$$
\vspace*{-7mm}
\caption{Implication graph for Example~\ref{ex:lin}. Decision literals are double boxed. Decision levels are separated by dashed lines.\label{fig:small-1uip}}
\vspace*{-7mm}
\end{figure}

\begin{example}
\label{ex:lin}
Consider a simple constraint problem with variables $x_1, x_2$,$x_3, x_4,
x_5, x_6$ with all initial domain $\{0,1,2,3,4,5,6,7\}$, 
and three constraints: $x_1 + 2x_2 + 3x_3 + 4x_4 + 4x_5 \leq 30$,
$x_4 \leq 4 \rightarrow x_6 = 1$, and 
$x_5 \leq 4 \rightarrow x_6 = 0$.
Suppose we make the decisions: $x_1 \geq 1$ (nothing propagates), 
$x_2 \geq 2$ (propagates $x_4 \leq 6$ and $x_5 \leq 6$), 
and $x_3 \geq 3$. This propagates $x_4 \leq 4$ and $x_5 \leq 4$, which in turn propagates
$x_6 = 1$ and causes the constraint $x_5 \leq 4 \rightarrow x_6 = 0$ to fail.
Figure~\ref{fig:small-1uip} 
shows the implication graph when
the conflict occurs. The double boxes indicate decision literals while the
dashed lines partition literals into decision levels. Dotted lines are
literals that are irrelevant to the failure.
To obtain the 1UIP nogood we
start with the conflict nogood $\lit{x_5 \leq 2} \wedge \lit{x_6 = 1} \rightarrow \false$
which contains every literal directly connected to the $\false$ conclusion. 
We have two literals from the last decision level ($\lit{x_5 \leq 4}$ and 
$\lit{x_6 = 1}$).  Since $\lit{x_6 = 1}$ 
was the last literal to be inferred of those two, we
resolve the current nogood with $expl(\lit{x_6 = 1}) = \lit{x_4 \leq 4}$ 
obtaining
$\lit{x_4 \leq 4} \wedge \lit{x_5 \leq 4} \rightarrow \false$. 
We still have two literals of the last decision level so we 
replace $\lit{x_5 \leq 4}$ by $expl(\lit{x_5 \leq 4}) = \lit{x_1 \geq 1} \wedge \lit{x_2 \geq 2}
\wedge \lit{x_3 \geq 3}$ obtaining
$\lit{x_1 \geq 1} \wedge \lit{x_2 \geq 2}
\wedge \lit{x_3 \geq 3} \wedge \lit{x_4 \leq 4} \rightarrow \false$.
We then replace $\lit{x_4 \leq 4}$ by $expl(\lit{x_4 \leq 4}) = \lit{x_1 \geq 1} \wedge \lit{x_2 \geq 2}
\wedge \lit{x_3 \geq 4}$ obtaining
$\lit{x_1 \geq 1} \wedge \lit{x_2 \geq 2} \wedge \lit{x_3 \geq 3} \rightarrow \false$.
This is the 1UIP nogood since it contains only one literal from level 3.
\hfill $\Box$
\end{example}

\section{Generality of Explanations}\label{sec:gen}

Since the nogoods derived by the resolution proof system are formed by resolving the explanations generated by the propagators, the more general the explanations are, the more general the nogood derived will be. Using better explanations means that for the same amount of search, we can derive stronger nogoods that prove that a greater part of the search space is failed. The following definitions allow us to compare and assess how good an explanation is:

\begin{definition}
\label{def:moregen}
Given two possible explanations $E \equiv l_1 \wedge \ldots \wedge l_n
\rightarrow p$ and $E' \equiv k_1 \wedge \ldots \wedge k_m \rightarrow p$
for the inference $p$, $E'$ is \emph{strictly more general} than $E$ iff:
$\wedge_{i=1}^n l_i \Rightarrow \wedge_{i=1}^m k_i$ and $\wedge_{i=1}^m k_i
\nRightarrow \wedge_{i=1}^n l_i$. \qed
\end{definition}

\begin{definition}
\label{def:maxgen}
An explanation $E \equiv l_1 \wedge \ldots \wedge l_n \rightarrow p$ for the
inference $p$ is \emph{maximally general} w.r.t. language of resolution $L$,
if there does not exist another explanation $E'$ in $L$ which is strictly
more general than $E$. \qed
\end{definition}

Note that maximally general explanations are not necessarily unique.

\begin{example}
\label{ex:gen-expl1}
Consider a linear constraint $x_1 + 2x_2 + 3x_3 + 4x_4 \leq 30$ and a current domain of $x_1 = 1, x_2 = 2, x_3 = 3$. The propagator can infer that $x_4 \leq 4$. 
There are many possible explanations. For example, $\lit{x_1 = 1} \wedge
\lit{x_2 = 2} \wedge \lit{x_3 = 3} \rightarrow \lit{x_4 \leq 4}$ is a
perfectly valid explanation. However, it is not very general. A strictly
more general explanation is $\lit{x_1 \geq 1} \wedge \lit{x_2 \geq 2} \wedge
\lit{x_3 \geq 3} \rightarrow \lit{x_4 \leq 4}$. However, this is still not
maximally general in the standard LCG language. For example, $\lit{x_2 \geq
  1} \wedge \lit{x_3 \geq 3} \rightarrow \lit{x_4 \leq 4}$ is a maximally
general explanation which is more general than the one before. Similarly,
$\lit{x_1 \geq 1} \wedge \lit{x_2 \geq 2} \wedge \lit{x_3 \geq 2}
\rightarrow \lit{x_4 \leq 4}$ is another maximally general explanation.
\hfill $\Box$
\end{example}

Clearly, a good starting point for making the resolution proof system stronger is to ensure that the LCG solver is using maximally general explanations, so that we are making the most out of the existing language.

\begin{definition}
\label{def:univmaxgen}
An explanation $E \equiv l_1 \wedge \ldots \wedge l_n \rightarrow p$ for the
inference $p$ is \emph{universally maximally general} if it is maximally
general w.r.t. to the universal language $L$ consisting of all possible
logical expressions.
\qed
\end{definition}

If the universally maximally general explanation for an inference cannot be expressed as a conjunction of literals in the existing language, then it is a good indication that a language extension may be useful for increasing the generality of the explanations for this constraint. 

\begin{example}
Consider the inference from Example~\ref{ex:gen-expl1}. The universally maximal general explanation is: $\lit{x_1 + 2x_2 + 3x_3 \geq 11} \rightarrow \lit{x_4 \leq 4}$, since $x_1 + 2x_2 + 3x_3 \geq 11$ is a necessary and sufficient condition on the domain for us to infer $x_4 \leq 4$ from $x_1 + 2x_2 + 3x_3 + 4x_4 \leq 30$. Clearly, there is no way that $\lit{x_1 + 2x_2 + 3x_3 \geq 11}$ can be expressed equivalently as a conjunction of equality, disequality or inequality literals on $x_1, x_2, x_3, x_4$. So a language extension may be useful here.
\hfill $\Box$
\end{example}

\section{Extending the Language}
\label{sec:ext-lang}

We now consider how we can extend the language of resolution to give more general explanations. We first give a simple motivating example.

\begin{example}
  Consider the 0-1 knapsack problem, given by $x_1, \ldots, x_n \in
  \{0,1\}$, $\sum_{i=1}^n w_i x_i \leq W$, $\sum_{i=1}^n p_i x_i \geq \obj$,
  where $\obj$ is to be maximized, $w_i$ represents the weights of item $i$,
  $W$ is the total capacity of the knapsack, and $p_i$ is the profit of item
  $i$. A normal CP solver will require $O(2^n)$ to solve this problem. Using
  an LCG solver does no better, because the size of the smallest proof of
  optimality using only equality, disequality and inequality literals on the
  $x_i$ is still exponential in $n$. On the other hand, suppose we
  introduced literals to represent partial sums of form: $\sum_{i=1}^{k} w_i
  x_i \geq W - w'$ and $\sum_{i=1}^{k} p_i x_i \leq \obj - p'$ where $k$,
  $w'$, $p'$ are arbitrary constants. Then, it becomes possible to prove
  optimality in $O(nWP)$, where $P = \sum_{i=1}^n p_i$. This is because it
  is now possible to express nogoods such as: $\lit{\sum_{i=1}^{k} w_i x_i
    \geq W - w'} \wedge \lit{\sum_{i=1}^{k} p_i x_i \leq \obj - p'}
  \rightarrow \false$ which represent that we have proved that given only a
  weight limit of $w'$ for the items $k+1$ to $n$, there is no way we can
  pick a subset of them such that their profit sum to at least $p'$.  
\ignore{If we
  modify the linear propagators in our LCG solver to use these new partial
  sum literals in their explanations and search the variables in the order
  $x_1, \ldots, x_n$, then each failure will give us a nogood of the above
  form. We need to derive at most $O(nWP)$ such nogoods to complete the
  proof, because $k$ needs to take at most $n$ values, $w'$ needs to take at
  most $W$ values, and $p'$ needs to take at most $P$ values, so the total
  number of search nodes required by this modified LCG solver is $O(nWP)$
  which is pseudo-polynomial complexity rather than exponential complexity.
}
If we modify our LCG solver to use these new partial sum literals
the amount of search required can be $O(nWP)$
  which is pseudo-polynomial rather than exponential
  complexity. \qed
\end{example}

When deciding on language extensions, there are two main factors we have to
consider:
\begin{itemize}
\item 
Are they going to improve the size of the resolution proof? 
\item 
What is the overhead of introducing the literal into the system? 
\end{itemize}
When we extend the language by introducing a literal $\lit{e}$ where $e$ is some logical expression over existing variables, we have to keep track of the truth value of $\lit{e}$ so that we can propagate any nogoods with this literal in it. This is accomplished by enforcing a channeling constraint $\lit{e} \leftrightarrow e$. For example, if we introduced a literal $\lit{x_1 + 2x_2 +3x_3 \geq 10}$, we would have to enforce the channeling constraint: $\lit{x_1 + 2x_2 + 3x_3 \geq 10} \leftrightarrow x_1 + 2x_2 + 3x_3 \geq 10$. Depending on what the expression is, this could be cheap or expensive.

As mentioned in the previous section, literals which allow global
propagators to explain their inferences in a more general way are prime
candidates for language extensions. Another benefit of such literals is that
they often represent intermediate logical concepts in the propagation
algorithm which the propagator is already keeping track of, and thus the
channelling propagation required to enforce $\lit{e} \leftrightarrow e$ can
be ``piggy-backed'' onto the original propagator at little extra cost. 

Ideally we can add a set of literals which allow the universally maximally general explanation for each inference to be described. Unfortunately, this is not always possible as the universally maximally general explanation may be some complicated logical expression that we cannot easily check the truth value of during search. Instead, we may have to settle for less general but more practical language extensions. We now analyze a number of global constraints to see what the maximally general explanations are given the standard LCG language consisting of equality, disequality and inequality literals on existing variables, and show the language extensions which can provide stronger explanations.

\subsection{Linear}
\label{subsec:lin-ext-res}

Linear constraints are by far the most common constraint appearing in
models.

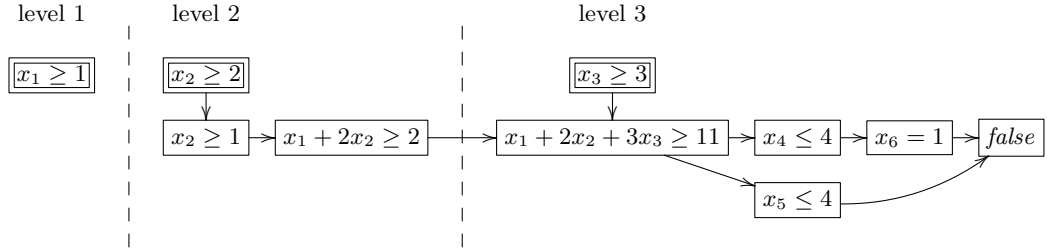
\begin{figure}[t]
\footnotesize 
\newcommand{\xyo}[1]{*+[F-]{#1}}
\newcommand{\xyd}[1]{*+[F=]{#1}} 
\newcommand{\xydd}[1]{*+[F.]{#1}}
$$
\hspace*{-10mm}
\xymatrix@=10pt{ 
   && \mbox{level 1} & \ar@{--}[dddd] & \mbox{level 2} && \ar@{--}[dddd]
& \mbox{level 3} \\
  && \xyd{x_1 \geq 1}
  &&   \xyd{x_2 \geq 2}
   \ar[d]
  &&& \xyd{x_3 \geq 3}
      \ar@{->}[d] 
 \\  
  &&&& \xyo{x_2 \geq 1} \ar[r] &\xyo{x_1 + 2x_2 \geq 2} \ar[rr]  && \xyo{x_1 + 2x_2 + 3x_3 \geq 11} \ar[r]\ar[dr]  & \xyo{x_4 \leq 4} \ar[r] & \xyo{x_6 = 1}\ar[r] & \xyo{\false} \\
  &&&&& &&& \xyo{x_5 \leq 4} \ar@/_10pt/[rru] &   \\
&&&&&&&&&&&&&&&&& \\
}
$$
\vspace*{-7mm}
\caption{Implication graph for Example~\ref{ex:lin} using a stronger
  language of learning. \label{fig:er}}
\end{figure}

\begin{example}\label{ex:linext}
Consider the linear constraint $x_1 + 2x_2 + 3x_3 + 4x_4 + 4x_5 \leq 30$
of Example~\ref{ex:lin}. Given $x_1 \geq 1, x_2 \geq 2, x_3 \geq 3$, we can
infer $x_4 \leq 4$. There are multiple possible maximally general
explanation in the standard LCG language, e.g., $\lit{x_2 \geq 1} \wedge
\lit{x_3 \geq 3} \rightarrow \lit{x_4 \leq 4}$ or $\lit{x_1 \geq 1} \wedge
\lit{x_2 \geq 2} \wedge \lit{x_3 \geq 2} \rightarrow \lit{x_4 \leq
  4}$. However, none of them are the most general explanation possible. 
If we extended the language with literals representing partial sums, we can
now use the universally maximally general explanation: $\lit{x_1 + 2x_2 +
  3x_3 \geq 11} \rightarrow \lit{x_4 \leq 4}$. 
Using this intermediate literal the implication graph for
Example~\ref{ex:lin}
changes to that shown in Figure~\ref{fig:er}.
Both $\lit{x_4 \leq 4}$ and $\lit{x_5 \leq 4}$ are explained by
$\lit{x_1 + 2x_2 + 3x_3 \geq 11}$.
The new 1UIP is simply $\lit{x_1 + 2x_2 + 3x_3 \geq 11} \rightarrow
\false$.
This is a much stronger nogood that will prune more of the search space.
\qed
\end{example}

The channelling propagation which enforces the consistency of a partial sum
literal and the variables in the partial sum must itself be explained, and
we can similarly use the partial sum literals to give more general
explanations. 

\begin{example}
Consider Example~\ref{ex:lin} again. If we only generate explanations on demand
during conflict analysis, we can introduce new partial sum literals to
explain other partial sum literals in a maximally general fashion.
When $x_4 \leq 4$ is inferred by the linear constraint $x_1 + 2x_2 + 3x_3 +
4x_4 + 4x_5 \leq 30$, the chain of
explanations going backwards from $\lit{x_4 \leq 4}$ would be: $\lit{x_1 +
  2x_2 + 3x_3 \geq 11} \rightarrow \lit{x_4 \leq 4}$, $\lit{x_1 + 2x_2 \geq
  2} \wedge \lit{x_3 \geq 3} \rightarrow \lit{x_1 + 2x_2 + 3x_3 \geq 11}$,
$\lit{x_2 \geq 1} \rightarrow
\lit{x_1 + 2x_2 \geq 2}$, $\lit{x_2 \geq 2} \rightarrow \lit{x_2 \geq 1}$.
The implication graph is shown in Figure~\ref{fig:er}.
\qed
\end{example}

There are also a significant number of global constraints which are composed
of linear constraints along with other primitive constraints like channeling
constraints (e.g., $among$, $at\_most$, $at\_least$, $sliding\_sum$, $gcc$, 
etc). These
can similarly benefit from partial sum literal language extensions on the
linears they are composed from.

\subsection{Lex}
\label{subsec:lex-ext-res}

Consider the global lexicographical constraint: $\mathit{lex\_less}([x_1, \ldots, x_n],
[y_1, \ldots, y_n])$ which constrains the sequence $x_1, \ldots, x_n$ to be
lexicographically less than $y_1, \ldots, y_n$, i.e.,: $x_1 < y_1 \vee (x_1
= y_1 \wedge x_2 < y_2) \vee \ldots \vee (x_1 = y_1 \wedge \ldots \wedge
x_{n-1} = y_{n-1} \wedge x_n < y_n)$. Consider a partial assignment $x_1 =
1, y_1 = 1, x_2 = 2, y_2 = 2, x_3 = 3$. From the constraint, we can infer
that $y_3 \geq 3$. The maximally general explanation in the standard LCG
language is: $\lit{x_1 = 1} \wedge \lit{y_1 = 1} \wedge \lit{x_2 = 2} \wedge
\lit{y_2 = 2} \wedge \lit{x_3 \geq 3} \rightarrow \lit{y_3 \geq
  3}$. However, if we extend the language with literals to represent things
such as: $x_i \geq y_i$, we can explain it using: $\lit{x_1 \geq y_1} \wedge
\lit{x_2 \geq y_2} \wedge \lit{x_3 \geq 3} \rightarrow \lit{y_3 \geq
  3}$. This second explanation is strictly more general and can produce a
more general nogood. For example, if in another branch, we had $x_1 = 2, y_1
= 2, x_2 = 1, y_2 = 1, x_3 = 3$, the first nogood cannot propagate since
$\lit{x_1 = 1}$ is not true, but the second one can since $\lit{x_1 \geq
  y_1}$, $\lit{x_2 \geq y_2}$ are true. The other mode of propagation for a
$\mathit{lex\_less}$ constraint can make use of new literals of the form:
$x_i > y_i$. For example, if $x_1 = 1, y_1 = 1, x_2 = 2, x_3 = 4, y_3 = 3$,
we can infer $y_3 \geq 3$ and explain it with: $\lit{x_1 \geq y_1} \wedge
\lit{x_3 > y_3} \wedge \lit{x_2 \geq 2} \rightarrow \lit{y_2 \geq 3}$. Thus
$\lit{x_i \geq y_i}$ and $\lit{x_i > y_i}$ are good language extension
candidates for $\mathit{lex\_less}$.

\subsection{Disjunctive}
\label{subsec:disj-ext-res}

Consider a disjunctive constraint $\disjunctive([s_1,s_2],[5,5])$ 
over two tasks with start times
having current domains of $s_1 \in \{2,\ldots,8\}, s_2 \in \{0, \ldots,
4\}$, and durations $d_1 = d_2 = 5$. A global propagator would reason that
task 1 must be scheduled after task 2, and therefore that $s_1 \geq
5$. There are multiple maximally general explanation in the standard LCG
language, e.g., $\lit{s_1 \geq 0} \wedge \lit{s_2 \leq 4} \wedge \lit{s_2
  \geq 0} \rightarrow \lit{s_1 \geq 5}$. We can 
extend the language with
literals to represent that task $i$ runs before task $j$, written as
$\lit{\sched{i}{j}}$ and channeled via: $\lit{\sched{i}{j}} \rightarrow s_i + d_i \leq s_j$, $\neg \lit{\sched{i}{j}} \rightarrow s_j + d_j \leq s_i$. Then we can explain the inference via: $\lit{\sched{2}{1}} \wedge
\lit{s_2 \geq 0} \rightarrow \lit{s_1 \geq 5}$. A nogood created from this
explanation have $\lit{\sched{2}{1}}$ 
rather than $\lit{s_1 \geq 0} \wedge
\lit{s_2 \leq 4}$ in it and will be more general. For example, if we have
another domain with $s_1 \geq 3, s_2 \leq 7, s_2 \geq 0$, a nogood created
from the first two explanations would not be able to propagate, but the
second one might since $s_1 \geq 3, s_2 \leq 7$ will cause 
$\lit{\sched{2}{1}}$ to
become true.

\subsection{Table}
\label{subsec:table-ext-res}

Consider a table constraint $\tablec([x_1, x_2, x_3, x_4]$, $[[1, 2, 3, 4]$, $[4, 3, 2, 1]$, $[1, 2, 2, 3]$, $[3, 1, 2, 1]$, $[1, 1, 1, 1]])$. Suppose we have $x_1 = 1$, $x_2 = 2$. Among other things, propagation will infer $x_4 \neq 1$. There are a number of different maximally general explanations in the standard LCG language, e.g., $\lit{x_1 \neq 4} \wedge \lit{x_1 \neq 3} \wedge \lit{x_2 \neq 1} \rightarrow \lit{x_4 \neq 4}$, or $\lit{x_2 \neq 3} \wedge \lit{x_2 \neq 1} \wedge \lit{x_2 \neq 1} \rightarrow \lit{x_4 \neq 4}$. However, none of these give the most general reason for $\lit{x_4 \neq 4}$. Suppose we extend the language with literals $r_i$ which represent whether the $i$th tuple is taken or not, i.e., $r_1 \equiv \lit{x_1 = 1 \wedge x_2 = 2 \wedge x_3 = 3 \wedge x_4 = 4}$, $r_2 \equiv \lit{x_4 = 1 \wedge x_2 = 3 \wedge x_3 = 2 \wedge x_4 = 1}$, etc. Then we can explain $\lit{x_4 \neq 4}$ using $\neg r_2 \wedge \neg r_4 \wedge \neg r_5 \rightarrow \lit{x_4 \neq 4}$. This is a universally maximally general explanation, i.e., if any domain knocks out tuples 2, 4 and 5 (which are the only ones that support $x_4 = 1$), then $x_4 \neq 1$.

Similarly, the explanations for the $\mathit{regular}$ constraint can be improved by introducing literals representing the intermediate states of the automata, and the explanations for binary/multi-decision diagram constraints (BDD/MDD) can be improved by introducing literals representing whether we take a particular node in the BDD/MDD or not. The global constraints $\alldiff$, $\circuit$, and many others also have language extensions that can give stronger explanations. However, the extensions we can think of are most likely impractical due to the expense of the channelling constraints.

\section{Exploiting Global Structure for Linears}\label{sec:struct}

For $\mathit{lex}$, $\mathit{table}$, $\mathit{disjunctive}$, $\mathit{regular}$ and $\mathit{bdd}$/$\mathit{mdd}$, the number of useful literals identified in Section~\ref{sec:ext-lang} is only linear or quadratic in the size of the constraint. Furthermore, all of those logical expressions are already things maintained internally by the global propagator and it is easy to alter the propagators to channel these literals and use them in explanations. Thus the overhead of adding these literals is fairly low and it is fine to simply add them all to the language. Linear on the other hand is more difficult. Linear constraints are extremely common and we know for certain that language extensions can be useful for this constraint. On the other hand, there are $O(adn2^n)$ possible partial sum literals for a length $n$ linear with largest coefficient $a$ and maximum domain size $d$. If we add too many of them, the cost of channeling them will swamp out any benefit we may get from search space reduction. In the worse case, we may have to calculate an exponential number of partial sums at each node just to channel them. We propose to only add the partial sum literals along a certain ordering of the terms in the linear, and to use the global structure of the problem in order to pick the ordering we use. 

Suppose that we had a particular ordering of the variables and suppose we
had a linear constraint $\sum_{i=1}^n a_i x_i \leq a_0$. Without loss of
generality assume that for each $i$, $x_i$ is before $x_{i+1}$ in our chosen
ordering (if not, just move the terms in the linear around and relabel the
indices). We propose to add only partial sum literals of the form
$\lit{\sum_{i=1}^k a_i x_i \geq v}$ 
for $1 \leq k < n$, e.g., $\lit{a_1 x_1 + a_2 x_2 \geq 3}$, but not $\lit{a_1 x_1 + a_3 x_3 \geq 3}$. The benefit here is that a single forward and backward pass through the terms is sufficient to channel the values of all these literals. On the forward pass, we aggregate the lower bound on $\sum_{i=1}^k a_i x_i$, where we start with $\sum_{i=1}^0 a_i x_i \geq 0$ and each subsequent term $\sum_{i=1}^k a_i x_i$ is greater than or equal to either the lower bound of the previous term $\sum_{i=1}^{k-1} a_i x_i$ plus the lower bound of $a_k x_k$, or to $v$ where $v$ is the largest value such that $\lit{\sum_{i=1}^k a_i x_i \geq v}$ is currently true. Similarly, on the backward pass, we aggregate the maximum value of $\sum_{i=1}^k a_i x_i$, where we start with $\sum_{i=1}^n a_i x_i \leq a_0$, and each subsequent term $\sum_{i=1}^k a_i x_i$ is less than or equal to either the upper bound of the previous term $\sum_{i=1}^{k+1} a_i x_i$ minus the lower bound of $a_k x_k$, or to $v-1$ where $v$ is the smallest value such that $\lit{\sum_{i=1}^k a_i x_i \geq v}$ is currently false. These allow us to fix any of the values of the partial sum literals which should be fixed, and we can propagate upper bounds on $a_k x_k$ using the difference between the lower bound of $\sum_{i=1}^k a_i x_i$ and the upper bound of $\sum_{i=1}^{k+1} a_i x_i$. 

These partial sum literals can be lazily introduced only as needed, i.e., when we need to use one of them in a nogood. Whenever the nogood database is cleaned to remove inactive nogoods, we also remove any partial sum literal that is no longer in any nogood. To reduce overhead even further, we can also only allow partial sum literals to be introduced at regular intervals. For example, if the interval was 5, we would only allow literals of the form $\lit{\sum_{i=1}^5 a_i x_i \geq v}$, $\lit{\sum_{i=1}^{10} a_i x_i \geq v}$, etc, to be introduced. We claim that this is often sufficient to get most of the benefit of the language extension. Thus we can trade off less overhead for a smaller reduction in proof size.

Now we need to pick the ordering that gives us the most useful partial sum literals. Many constraint problems have structures such that each variable is only strongly related to a small subset of other variables. For example, in a disjunctive scheduling problem, tasks which have overlapping time intervals may be strongly related, while tasks whose time intervals are far apart may be weakly related. Or in a graph colouring problem, adjacent nodes are strongly related, but nodes far apart in the graph are weakly related. Such structure can be exploited in order to give smaller resolution proofs. A good search strategy will label variables which are strongly related to those already fixed, rather than to pick some completely random variable to label. This improves propagation and also allows stronger nogoods to be derived. It is well known that techniques such as caching~\cite{smith05}, variable elimination~\cite{larrosa2003}, dynamic programming~\cite{bellman}, and nogood learning allows a problem to be solved with a complexity that is only exponential in the width of the search order (assuming sufficient memory). We claim that the ordering which minimizes this width is also the ideal ordering to use in order to introduce the partial sum literals as it provides the most generality to the nogoods.

\begin{example}
Consider a simple problem with variables $x_1, \ldots, x_{10} \in \{1, \ldots, 5\}$, constraints $\mathit{all\_diff}(x_i, x_{i+1}, x_{i+2})$ for $i = 1, \ldots, 8$ and objective function $\sum x_i$ to be minimized. Given that the $\mathit{all\_diff}$ constraints constrain sets of consectutive variables, an ordering which minimizes the width is to label $x_1, \ldots, x_{10}$ in order. Suppose we are trying to find a solution with objective $\leq 21$. Suppose we made the decisions $x_1 = 1, x_2 = 2, x_3 = 3, x_4 = 4, x_5 = 5$ in that order. Propagation forces $x_6, \ldots, x_{10} \leq 2$, causing the $\mathit{all\_diff}$'s to fail. If we used the ordering $x_1, \ldots, x_{10}$ to introduce partial sum literals, the 1UIP nogood would be: $\lit{\sum_{i=1}^5 x_i \geq 15} \rightarrow \false$. If we used the ordering $x_1, x_3, x_5, x_7, x_9, x_2, x_4, x_6, x_8, x_{10}$ to introduce partial sum literals instead, the 1UIP would be: $\lit{x_2 \geq 2} \wedge \lit{x_4 \geq 4} \wedge \lit{x_1 + x_3 + x_5 \geq 9} \rightarrow \false$, which is far less general. \qed
\end{example}

In our experiments, we manually find a low width ordering of the variables. However, it is easy to automate this by using an approximate algorithm for calculating the pathwidth of the constraint graph (e.g.,~\cite{bodlaender95}) to give a good variable ordering.

\section{Experiments}
\label{sec:exp}

We perform 4 sets of experiments. \short{The first tests the effect that the various language extensions have on the number of nodes and time required to solve an instance. The second tests the tradeoff between the reduction in node count and the increase in overhead when introducing partial sum literals. The third tests different stategies for introducing partial sum literals. The fourth tests whether partial sum literals also improve proof size for dynamic search strategies.

}The experiments were performed on Xeon Pro 2.4GHz processors using the
state of the art LCG solver \chuffed{}. We use 8 problems. For brevity 
we only describe the global constraints and the structural order we
used to create partial sum literals in each problem with linear
constraints. MiniZinc models of these problems can be found at \url{www.cs.mu.oz.au/~pjs/ext-res/}. The knapsack problem has linear constraints. We pick an ordering which sorts the items such that the profit to weight ratio is descending. The concert hall problem~\cite{law06} and talent scheduling problem~\cite{banda11} are scheduling problems with linear constraints. We pick an ordering based on time from earliest to latest. The maximum density still life problem (CSPLib prob032) is a board type problem with linear constraints. We pick an ordering which goes row by row from top to bottom and left to right. The PC-board problem~\cite{Martin2005} and the balanced incomplete block design problem (BIBD)~\cite{meseguer99} are matrix problems with linear constraints. BIBD also has $\mathit{lex\_lesseq}$ symmetry breaking constraints. We use an ordering which goes row by row from top to bottom. The nonograms problem~\cite{Yu11} is a board type problem with $\mathit{regular}$ constraints. The jobshop scheduling problem~\cite{garey76} has $\mathit{disjunctive}$ constraints. We use a timeout of 600 seconds. In each table, \textit{fails} is the geometric mean of the the number of fails in the search, \textit{time} is the geometric mean of the time of search in seconds (with timeouts counting as 600), and \textit{svd} is the number of instances solved to optimality. Note that we use propagators of the exact same propagation strength in all the methods for all experiments, so any difference is purely due to nogood learning. 

The first experiment compares a CP solver without nogood learning ({\sf no-ngl}) with nogood learning using the basic language of equality, disequality and inequality literals on existing variables ({\sf basic-ngl}), and nogood learning with the language extensions described in Section~\ref{sec:ext-lang} ({\sf er-ngl}). For the 6 problems with linear constraints, we use a fixed order search based on the structural ordering we described above. We will test the same 6 problems with a dynamic search in the fourth experiment. For nonograms and jobshop, we use the weighted degree search heuristic~\cite{domwdeg}, which works well for these two problems. The results are shown in Table~\ref{tab:ext-res}. Clearly, we can get very significant reductions in node counts on a wide variety of problems. However, depending on how large the node reduction is and the overhead of extending the language, we may not always get a speedup (e.g., BIBD). The node reduction tends to grow exponentially with problem size.

\begin{table}
\footnotesize
\caption{\label{tab:ext-res} Comparison of the solver without nogood learning ({\sf no-ngl}) with nogood learning using the basic language of equality, disequality and inequality literals on existing variables ({\sf basic-ngl}), and nogood learning with the language extensions described in Section~\ref{sec:ext-lang} ({\sf er-ngl}).}
\begin{center}
\begin{tabular}{l|rrr|rrr|rrr}
Problem & \multicolumn{3}{c|}{\sf no-ngl} & \multicolumn{3}{c|}{\sf basic-ngl} & \multicolumn{3}{c}{\sf er-ngl} \\
   & \it fails & \it time & \it svd & \it fails & \it time & \it svd & \it
   fails & \it time & \it svd \\
\hline
Knapsack-30 & 24712 & 0.10 & 20 & 24526 & 0.55 & 20 & \bf 207 & \bf 0.04 & \bf 20 \\
Knapsack-40 & 2549810 & 7.91 & 20 & 2548993 & 68.14 & 20 & \bf 685 & \bf 0.17 & \bf 20 \\
Knapsack-100 & 119925896 & 600 & 0 & 10855544 & 600 & 0 & \bf 12100 & \bf 32.44 & \bf 20 \\
Concert-Hall-35 & 93231 & 2.95 & 20 & 22450 & 1.62 & 20 & \bf 1529 & \bf 0.31 & \bf 20 \\
Concert-Hall-40 & 1814248 & 63.12 & 18 & 389799 & 35.14 & 20 & \bf 8751 & \bf 2.45 & \bf 20 \\
Concert-Hall-45 & 10186173 & 375.1 & 7 & 3187357 & 329.7 & 9 & \bf 34433 & \bf 14.95 & \bf 20 \\
Talent-14 & 81220 & 2.33 & 20 & 17543 & 1.17 & 20 & \bf 6362 & \bf 0.66 & \bf 20 \\
Talent-16 & 572341 & 17.67 & 20 & 111403 & 11.12 & 20 & \bf 20629 & \bf 2.98 & \bf 20 \\
Talent-18 & 8369293 & 256.3 & 16 & 1535814 & 215.1 & 16 & \bf 89813 & \bf 20.81 & \bf 20 \\
Still-Life-9 & 726722 & 39.42 & 1 & 123544 & 13.43 & 1 & \bf 13687 & \bf 3.07 & \bf 1 \\
Still-Life-10 & 3390292 & 189.25 & 1 & 478182 & 57.14 & 1 & \bf 10165 & \bf 2.39 & \bf 1 \\
Still-Life-11 & 9727533 & 600 & 0 & 4329170 & 600 & 0 & \bf 76225 & \bf 37.44 & \bf 1 \\
PC-Board & 16944535 & 405.8 & 21 & 68946 & 7.35 & 97 & \bf 34064 & \bf 6.66 & \bf 97 \\
BIBD & 4660894 & 126.3 & 7 & 125689 & \bf 25.66 & 13 & \bf 32588 & 28.57 & \bf 15 \\
Nonogram-small & 13566 & 8.02 & 11 & 1827 & 2.17 & 11 & \bf 854 & \bf 1.13 & \bf 11 \\
Nonogram-medium & 299092 & 254.2 & 3 & 2822 & 5.47 & 4 & \bf 1491 & \bf 2.97 & \bf 4 \\
Nonogram-large & 829090 & 600 & 0 & 51449 & 73.18 & 4 & \bf 15166 & \bf 27.61 & \bf 4 \\
Jobshop-8 & 16459 & 0.65 & 20 & 489 & 0.08 & 20 & \bf 357 & \bf 0.06 & \bf 20 \\
Jobshop-10 & 2260393 & 167.6 & 13 & 6266 & 1.74 & 20 & \bf 3404 & \bf 1.02 & \bf 20 \\
Jobshop-12 & 6848535 & 596.1 & 1 & 87619 & 41.93 & 20 & \bf 40825 & \bf 18.43 & \bf 20 \\
\end{tabular}
\end{center}
\vspace*{-10mm}
\end{table}

In the second set of experiments, we test what happens if we only introduce partial sum literals every 5, 10, 20 or 50 variables as described in Section~\ref{sec:struct}. For ease of comparison, we also repeat the {\sf er-ngl} column from above, where we introduced partial sum literals after every variable. The results are shown in Table~\ref{tab:split-len}. The trend is very clear here. The fewer partial sum literals we add, the less reduction in node count we have. However, it also requires less overhead. For many of the instances, introducing partial sum literals every 5 to 10 variables is optimal. 

\begin{table}
\footnotesize
\caption{\label{tab:split-len} Comparison of introducing partial sum literals after every 1, 5, 10, 20 or 50 terms in the linear constraints.}
\vspace*{-5mm}
\begin{center}
\begin{tabular}{l|rr|rr|rr|rr|rr}
Problem & \multicolumn{2}{c|}{\sf 1} & \multicolumn{2}{c|}{\sf 5} & \multicolumn{2}{c|}{\sf 10} & \multicolumn{2}{c|}{\sf 20} & \multicolumn{2}{c}{\sf 50} \\
   & \it fails & \it time & \it fails & \it time & \it fails & \it time & \it fails & \it time & \it fails & \it time  \\
\hline
Knapsack-30 & \bf 207 & 0.04 & 461 & \bf 0.02 & 1157 & 0.05 & 12301 & 0.37 & 24526 & 0.56 \\
Knapsack-40 & \bf 685 & 0.17 & 1957 & \bf 0.14 & 8961 & 0.41 & 86123 & 3.24 & 2548994 & 67.07 \\
Knapsack-100 & \bf 12100 & 32.44 & 46677 & 42.87 & 387626 & 166.02 & 4689069 & 600 & 12267788 & 600 \\
Concert-Hall-35 & \bf 1529 & 0.31 & 1973 & \bf 0.20 & 3042 & 0.26 & 4450 & 0.34 & 22450 & 1.62 \\
Concert-Hall-40 & \bf 8751 & 2.45 & 11125 & \bf 1.37 & 16703 & 1.75 & 29582 & 2.97 & 389800 & 34.65 \\
Concert-Hall-45 & \bf 34433 & 14.95 & 44658 & \bf 7.78 & 66910 & 10.69 & 178391 & 30.19 & 3196963 & 329.77 \\
Talent-14 & \bf 6362 & \bf 0.66 & 9196 & 0.67 & 17543 & 1.19 & 17543 & 1.19 & 17543 & 1.20 \\
Talent-16 & \bf 20629 & \bf 2.98 & 31906 & 3.27 & 111403 & 11.02 & 111403 & 11.08 & 111403 & 10.96 \\
Talent-18 & \bf 89813 & 20.81 & 135955 & \bf 20.26 & 1534795 & 215.7 & 1535554 & 216.1 & 1532865 & 216.5 \\
Still-Life-9 & \bf 13687 & 3.07 & 13952 & 2.07 & 14979 & \bf 2.04 & 20556 & 2.62 & 30752 & 3.89 \\
Still-Life-10 & 10165 & 2.39 & \bf 9252 & \bf 1.37 & 11090 & 1.46 & 13301 & 1.64 & 23797 & 2.84 \\
Still-Life-11 & 76225 & 37.44 & \bf 70162 & \bf 22.08 & 106333 & 40.01 & 94390 & 28.19 & 251276 & 77.91 \\
PC-Board & \bf 34064 & 6.66 & 46387 & \bf 6.32 & 68941 & 7.39 & 68943 & 7.41 & 68948 & 7.41 \\
BIBD & \bf 32588 & 28.57 & 45998 & 15.62 & 55281 & \bf 14.63 & 94860 & 22.75 & 96483 & 18.88 \\
\end{tabular}
\end{center}
\vspace*{-5mm}
\end{table}

We now compare different ways of picking the order for creating partial sum literals. We compare the structure based ordering ({\sf struct}) which we used in the previous experiments, a random ordering ({\sf random}), and an ordering based on sorting on the size of the coefficients in descending order ({\sf coeff}). We also repeat the no language extension column {\sf basic} and the {\sf er-ngl}  (renamed to {\sf struct}) from the first experiment for ease of comparison. The results are shown in Table~\ref{tab:order}. It can be seen that even when we use an ordering which is inconsistent with the structure of the problem, we can still get some reduction in node count. However, the much smaller reduction in node count means that the overhead may often swamp out any benefit from the reduced search. A random ordering generally gives the least reduction in node count out of the three and the structure based ordering generally gives the most. An ordering based on the size of the coefficients is somewhere in the middle, depending on whether the coefficients happen to follow the structure of the problem or not.

\begin{table}
\footnotesize
\caption{\label{tab:order} Comparison of different orderings for generating partial sum literals.}
\begin{center}
\begin{tabular}{l|rr|rr|rr|rr}
Problem & \multicolumn{2}{c|}{\sf basic} & \multicolumn{2}{c|}{\sf struct} & \multicolumn{2}{c|}{\sf random}& \multicolumn{2}{c}{\sf coeff} \\
   & \it fails & \it time & \it fails & \it time & \it fails & \it time & \it fails & \it time \\
\hline
Knapsack-30 & 24526 & 0.55 & \bf 207 & \bf 0.04 & 15873 & 4.02  & 9882 & 2.47 \\
Knapsack-40 & 2548993 & 68.14 & \bf 685 & \bf 0.17 &  849769 & 350.1 & 424910 & 196.4 \\
Knapsack-100 & 10855544 & 600 & \bf 12100 & \bf 32.44 & 153826 & 600 & 130592 & 600 \\
Concert-Hall-35 & 22450 & 1.62 & \bf 1529 & \bf 0.31 & 16086 & 4.12 & 13490 & 3.43 \\
Concert-Hall-40 & 389799 & 35.14 & \bf 8751 & \bf 2.45 & 217273 & 77.46 & 179433 & 62.65 \\
Concert-Hall-45 & 3187357 & 329.7 & \bf 34433 & \bf 14.95 & 1021842 & 438.8 & 896397 & 395.3 \\
Talent-14 & 17542 & 1.17 & \bf 6362 & \bf 0.66 & 9877 & 1.16 & 7345 & 0.78 \\
Talent-16 & 111403 & 11.12 & \bf 20629 & \bf 2.98 & 40954 & 6.70 & 26212 & 3.71 \\
Talent-18 & 1535814 & 215.1 & \bf 89813 & \bf 20.81 & 277154 & 63.06 & 137521 & 29.40 \\
Still-Life-9 & 123544 & 13.43 & \bf 13687 & \bf 3.07 & 120151 & 38.33 & 13687 & 3.09 \\
Still-Life-10 & 478182 & 57.14 & \bf 10165 & \bf 2.39 & 167449 & 59.63 & 10165 & 2.39 \\
Still-Life-11 & 4329170 & 600 & \bf 76225 & \bf 37.44 & 1251142 & 600 & 76225 & 37.44 \\
PC-Board & 68946 & 7.35 & 34064 & \bf 6.66 & 52425 & 9.00 & \bf 34054 & 6.66 \\
BIBD & 125689 & \bf 25.66 & \bf 32588 & 28.57 & 38233 & 36.41 & 32632 & 28.62 \\
\end{tabular}
\end{center}
\vspace*{-7mm}
\end{table}

Finally, we test whether the node reduction we gain from partial sum literals is dependent on a specific search order, or whether we will benefit even if we use a dynamic search strategy. We use the variable state independent decaying sum (VSIDS) heuristic~\cite{moskewicz01} adapted from SAT. This is the standard search heuristic used in most current state of the art SAT solvers and is also very effective for some CP problems. For easy comparison, we also repeat the column {\sf er-ngl} from Table~\ref{tab:ext-res}, which we rename to {\sf er-ngl-fixed}. We call VSIDS on the basic language {\sf basic-ngl-vsids} and on the extended language {\sf er-ngl-vsids}. The results are shown in Table~\ref{tab:vsids}. It can be seen that extending the language reduces the node count on all the problems tested. However, VSIDS is not as capable of exploiting the new literals as a fixed order search using the structure based ordering in knapsack, concert hall, talent scheduling or maximum density still life. On PC board and BIBD, VSIDS is far superior to the fixed order search even without any language extension. The language extension does result in a reduced node count, but the overhead swamps out any benefits.

\begin{table}
\footnotesize
\caption{\label{tab:vsids} Comparison between VSIDS search heuristic on the basic language and the extended language.}
\begin{center}
\begin{tabular}{l|rrr|rrr|rrr}
Problem & \multicolumn{3}{c|}{\sf er-ngl-fixed} & \multicolumn{3}{c|}{\sf basic-ngl-vsids} & \multicolumn{3}{c}{\sf er-ngl-vsids} \\
   & \it fails & \it time & \it svd & \it fails & \it time & \it svd & \it fails & \it time & \it svd \\
\hline
Knapsack-30 & \bf 207 & \bf 0.04 & \bf 20 & 6562 & 0.11 & 20 & 1951 & 0.54 & 20 \\
Knapsack-40 & \bf 685 & \bf 0.17 & \bf 20 & 179932 & 5.71 & 20 & 18821 & 5.89 & 20 \\
Knapsack-100 & \bf 12100 & \bf 32.44 & \bf 20 & 7017850 & 600 & 0 & 140533 & 600 & 0 \\
Concert-Hall-35 & \bf 1529 & \bf 0.31 & \bf 20 & 84535 & 6.76 & 20 & 17165 & 4.74 & 20 \\
Concert-Hall-40 & \bf 8751 & \bf 2.45 & \bf 20 & 1703836 & 156.94 & 18 & 113479 & 45.60 & 20 \\
Concert-Hall-45 & \bf 34433 & \bf 14.95 & \bf 20 & 5087797 & 510.95 & 4 & 502381 & 280.84 & 15 \\
Talent-14 & \bf 6362 & \bf 0.66 & \bf 20 & 21981 & 1.41 & 20 & 10904 & 2.57 & 20 \\
Talent-16 & \bf 20629 & \bf 2.98 & \bf 20 & 139837 & 8.86 & 20 & 29529 & 6.61 & 20 \\
Talent-18 & \bf 89813 & \bf 20.81 & \bf 20 & 632778 & 45.74 & 19 & 202930 & 46.83 & 18 \\
Still-Life-9 & \bf 13687 & \bf 3.07 & \bf 1 & 2412735 & 318.1 & 1 & 46829 & 12.96 & 1 \\
Still-Life-10 & \bf 10165 & \bf 2.39 & \bf 1 & 3813974 & 600 & 0 & 86837 & 27.17 & 1 \\
Still-Life-11 & \bf 76225 & \bf 37.44 & \bf 1 & 4341884 & 600 & 0 & 195013 & 81.6 & 1 \\
PC-Board & 34064 & 6.66 & 97 & 11803 & \bf 0.99 & \bf 100 & \bf 11576 & 1.73 & 100 \\
BIBD & 32588 & 28.57 & 15 & 3489 & \bf 0.38 & \bf 28 & \bf 2207 & 1.21 & 28 \\
\end{tabular}
\end{center}
\vspace*{-7mm}
\end{table}

\section{Related Work}\label{sec:related}

The work presented here is closely related to the body of work on Boolean encodings for global constraints in SAT. Better Boolean encodings can be smaller in size and can also improve the power of the proof system, leading to faster solves (e.g.,~\cite{sinz05,bailleux03,een06}). However, our approach has several advantages. When developing Boolean encodings for global constraints for use in a SAT solver, the primary concerns are: 1) the size of the encoding, and 2) its propagation strength. All of the literals and clauses in the encoding must be statically created in the SAT solver. As a result, the encoding has to be reasonably small or the SAT solver will run out of memory or slow to a crawl. These concerns are far less important in the context of LCG. This is because an LCG solver does not ever need to produce a static Boolean encoding of the global constraint. Instead, the high level CP global propagator is fully responsible for all propagation (so we always get the full propagation strength), and it lazily creates literals and clauses as needed. As a result, even if there are potentially an exponential number of possible literals and clauses, it is typically still fine, because during any one solve, only a very small proportion of those literals and clauses will need to be created, and they can be thrown away as soon as they are no longer useful~\cite{feydy09}. This flexibility means that we are able to consider other important factors such as which literals will improve the power of the proof system. For example, when a SAT solver encodes a pseudo Boolean constraint into clauses via a BDD translation, the variable order must be chosen to make the BDD small, which may not be ideal for the power of the proof system. On the other hand, we can pick an order which is better for the power of the proof system even if the potential number of literals introduced is very large.

The closest related work to that presented here is \emph{conflict directed
  lazy decomposition}~\cite{cp2012a}. Lazy decomposition treats a global
propagator for $c$ as a black box that hides a SAT encoding that implements
$c$. As computation progresses the propagator for $c$ lazily exposes more
and more of this SAT encoding if an activity heuristic indicates that this
may be beneficial. \short{In effect, it propagates the constraint half with
  the visible part of the SAT encoding and half with the global
  propagator. Lazy decomposition extends the language of resolution by
  exposing literals in the SAT encoding, hence it is a structure based
  approach to extended resolution.} These exposed Boolean variables give us
a structure based extension to the language. However, lazy decomposition is
complex to apply, as one must be able to effectively split a propagator into
two parts. The only constraints for which lazy decomposition is defined
in~\cite{cp2012a} are cardinality and pseudo-Boolean constraints. Lazy
decomposition has the advantage that it uses the activity of
literals in the search as a heuristic to determine whether adding the
partial sum literals will be useful. It has the disadvantage that it does
not use the global structure to determine which intermediate literals to
add. Instead, for psuedo-Boolean constraints it uses the size of the coefficients to order the partial
sums, as this tends to reduce the size of the Boolean encoding. However, as
Table~\ref{tab:order} shows, using a ordering different from the structural
one can nullify most of the search space reduction. \short{Lazy
  decomposition is also not as lazy as its name suggests. When this approach
  does decide to introduce partial sum literals for a particular partial
  sum, it immediately introduces them for all possible bounds (which can be
  exponentially many), along with all the clauses linking them. Our approach
  only introduces the literals that are needed for explanation, and they can
  be cleaned up when they are no longer needed.}

\short{There have been several works on using extended resolution in SAT solving. In~\cite{?,?}, the authors propose a method of BDD-based symbolic SAT solving which can be seen as a form of extended resolution. However, this method is only effective on certain classes of problems and does not generalize to the class of conflict directed clause learning (CDCL) solvers which are the current state of the art in SAT solving. In~\cite{?}, the authors propose to use a form of extended resolution for symmetry breaking. However, this approach is specific to the graph colouring problem.} There have been several works on using extended resolution in clause learning SAT solvers. In~\cite{audemard10}, the extension considered are of the form: if $\bar{l_1} \vee \alpha$ and $\bar{\l_2} \vee \alpha$ are two successively derived nogoods, then add the new literal $z$ defined via $z \leftrightarrow l_1 \vee l_2$. In~\cite{huang10}, another extension rule is proposed where they add a new literal $z \leftrightarrow \bar{d_1} \vee \ldots \vee \bar{d_k}$ where the $d_i$ are a subset of the assignments which led to a conflict. While these extensions appears useful for some SAT instances, it seems unlikely that these methods will be able to generate the right extensions for CP problems. For example, a partial sum literal such as $\lit{x_1 + x_2 \geq 5}$ is defined by $\lit{x_1 + x_2 \geq 5} \leftrightarrow \ldots \vee (\lit{x_1 \geq 0} \wedge \lit{x_2 \geq 5}) \vee (\lit{x_1 \geq 1} \wedge \lit{x_2 \geq 4}) \vee \ldots$. Considering the size of the definition of a general partial sum literal, it will take an incredible amount of luck for one of the above methods to introduce a literal that means exactly a partial sum literal. It is far more effective to keep the structural information contained in a high level model and to use that to introduce useful literals. Once a problem has been converted into conjunctive normal form, so much of the structural information has been lost that it is very difficult for any automated methods to be able to ``rediscover'' the literals that matter.

\section{Conclusion}\label{sec:conc}

\ignore{
Extending the language of resolution can make an exponential difference in the size of a proof of unsatisfiability or optimality. By analyzing the global constraints of the problem we can determine language extensions that are likely to reduce the
size of the proof, and we can extend the global propagator implementation to
enforce the relationship between these new literals and other literals.
The experiments show that extended resolution based on structure can be
highly beneficial in solving combinatorial optimization problems. 
}
A significant amount of research in CP has focused on improving the power of the proof systems used in CP solvers by developing more powerful global propagators. However, such research may well be nearing their limits as the optimal propagators for most commonly used constraints are already known. Nogood learning provides an orthogonal way in which to improve the power of the proof systems used in CP solvers. Extending the language of resolution changes the power of the resolution proof system used in nogood learning and can exponentially reduce the size of a proof of unsatisfiability or optimality, leading to much faster CP solving. The primary difficulty of using extended resolution is in finding the right language extensions to make. We have given a framework for analyzing the generality of explanations made by global propagators in LCG solvers, and shown that language extensions which improve the generality of these explanations are excellent candidates for language extension. Experiments show that such structure based extended resolution can be highly beneficial in solving a wide range of combinatorial optimization problems.

\ignore{
\subsubsection*{Acknowledgments}
NICTA is funded by the Australian Government as represented 
by the Department of Broadband, 
Communications and the Digital Economy and the Australian Research Council.
}

\bibliography{paper}
\bibliographystyle{plain}

\end{document}